\documentclass[letterpaper,conference,10pt]{ieeeconf}
\IEEEoverridecommandlockouts
\usepackage{amsmath,graphicx,epsfig,color,amsfonts}

\graphicspath{{./fig/}}

\usepackage{verbatim}
\usepackage{amssymb}
\usepackage{mathtools}
\usepackage{commath}
\usepackage{nth}
\usepackage{cite}
\usepackage{cleveref}
\usepackage{caption}
\usepackage{subcaption}
\usepackage[vlined,ruled]{algorithm2e}

\def\expt{\mathbb{E}}
\def\real{\mathbb{R}}

\newcommand{\until}[1]{\{1,\dots, #1\}}
\newcommand{\subscr}[2]{#1_{\textup{#2}}}
\newcommand{\supscr}[2]{#1^{\textup{#2}}}
\newcommand{\setdef}[2]{\{#1 \; | \; #2\}}
\newcommand{\seqdef}[2]{\{#1\}_{#2}}

\newcommand{\ceil}[1]{\left\lceil #1 \right\rceil}

\DeclareMathOperator*{\argmax}{arg\,max}
\DeclareMathOperator*{\argmin}{arg\,min}

\newcommand\oprocendsymbol{\hbox{$\square$}}
\newcommand\oprocend{\relax\ifmmode\else\unskip\hfill\fi\oprocendsymbol}

\renewcommand{\baselinestretch}{1}

\def \bs {\boldsymbol}
\def \mc {\mathcal}

\def \etal {\emph{et al.}}

\newtheorem{theorem}{Theorem}

\newtheorem{lemma}[theorem]{Lemma}

\newtheorem{remark}{Remark}

\newtheorem{assumption}{Assumption}
\newtheorem{definition}{Definition}

\renewcommand{\theenumi}{(\roman{enumi}}

\usepackage{caption}
\usepackage{balance}
\title{\bf
Multi-Robot Gaussian Process Estimation and Coverage: \\ A Deterministic Sequencing Algorithm and Regret Analysis

\thanks{This work has been supported in part by NSF grant IIS-1734272 and ARO grant W911NF-18-1-0325.}
}

\author{Lai Wei \hspace{0.5in} Andrew McDonald \hspace{0.5in} Vaibhav Srivastava
\thanks{L. Wei and V. Srivastava are with the Department of Electrical and Computer Engineering. Michigan State University, East Lansing, MI 48823 USA. {\tt\small e-mail: \{weilai1, vaibhav\}@msu.edu}} 
\thanks{A. McDonald
is with the Department of Computer Science and Engineering. Michigan State University, East Lansing, MI 48823 USA.
{\tt\small e-mail: mcdon499@msu.edu}}%
}

\begin{document}

\maketitle
\thispagestyle{empty}
\pagestyle{empty}

\newcommand{\rtwo}{\mathbb{R}^2}
\newcommand{\rone}{\mathbb{R}}
\newcommand{\qone}{\mathbb{Q}}
\newcommand{\qtwo}{\mathbb{Q}^2}
\newcommand{\nat}{\mathbb{N}}
\newcommand{\ex}{\mathbb{E}}
\newcommand{\rt}{\rightarrow}
\newcommand{\lt}{\leftarrow}
\newcommand{\T}{\top}
\newcommand{\normal}{\mathcal{N}}
\newcommand{\vor}{\mathcal{V}}
\newcommand{\st}{\mid}
\renewcommand{\abs}[1]{\left|#1\right|}
\renewcommand{\set}[1]{\left\{#1\right\}}
\renewcommand{\norm}[1]{\left\lVert #1\right\rVert}
\newcommand{\obar}[1]{\overline{#1}}
\newcommand{\cov}{\text{cov}}

\newcommand{\bvec}[1]{\bs{#1}}


\begin{abstract}%
\label{sec:abstract}%
We study the problem of distributed multi-robot coverage over an unknown, nonuniform sensory field. Modeling the sensory field as a realization of a Gaussian Process and using Bayesian techniques, we devise a policy which aims to balance the tradeoff between \emph{learning} the sensory function and \emph{covering} the environment. We propose an adaptive coverage algorithm called Deterministic Sequencing of Learning and Coverage (DSLC) that schedules learning and coverage epochs such that its emphasis gradually shifts from exploration to exploitation while never fully ceasing to learn. Using a novel definition of coverage regret which characterizes overall coverage performance of a multi-robot team over a time horizon $T$, we analyze DSLC to provide an upper bound on expected cumulative coverage regret. Finally, we illustrate the empirical performance of the algorithm through simulations of the coverage task over an unknown distribution of wildfires. 
\end{abstract}


\section{Introduction}
\label{sec:introduction}
Autonomous systems must remain robust and resilient in the face of uncertainty, capable of making decisions under the influence of imperfect and incomplete information. Real-world environments are unpredictable, noisy, and stochastic by their nature---various factors including weather, terrain, and human behavior combine with changing mission goals and operating constraints to necessitate adaptive policies. In order to successfully deal with uncertainty, autonomous systems must strike a balance between exploration and exploitation, simultaneously learning about their environment while accomplishing a task that depends on their collective knowledge of it. 


The \emph{coverage problem}~\cite{Cortes2004} arises naturally in multi-robot systems when a team of agents wishes to deploy themselves over an environment according to a particular sensory function $\phi$, which specifies the degree to which a robot is ``needed.'' Equivalently, the team of agents aims to partition an environment and achieve a configuration which minimizes the coverage cost defined by the sum of the $\phi$-weighted distances from every point in the environment to the nearest agent. Example applications of coverage range from search and rescue to wildfire fighting, smart agriculture, ecological surveying, environmental cleanup, and climate monitoring.

In this paper, we study the explore-exploit trade-off in \emph{adaptive coverage control}, wherein agents learn the sensory function $\phi$ while deploying themselves in the environment such that coverage cost is minimized, and derive an upper bound on the expected cumulative regret of the proposed policy. 



Classical approaches to coverage control \cite{Cortes2004, Cortes2005, Lekien2010, hussein2007effective} assume \emph{a priori} knowledge of $\phi$ and employ Lloyd's algorithm \cite{Lloyd1982} to guarantee the convergence of agents to a local minimum of the coverage cost. In these algorithms, each agent communicates with the agents in the neighboring partitions at each time and updates its partition. 
Distributed \emph{gossip-based} coverage algorithms \cite{Bullo2012} address potential communication bottleneck in classical approaches by updating partitions pairwise  between the agents in neighboring partitions. 

While much of the work in coverage considers continuous convex environments, a discrete graph representation of the environment is considered in~\cite{Durham2012}, which allows for non-convex environments. Additionally, gossip-based coverage algorithms in graph environments converge almost surely to pairwise-optimal partitions in finite time~\cite{Durham2012}.




Recent works have focused on the problem of \emph{adaptive} coverage, in which agents are not assumed to have knowledge of $\phi$ \emph{a priori}. Parametric estimation approaches to adaptive coverage  \cite{Schwager2009, Schwager2017} model $\phi$ as a linear combination of basis functions and propose algorithms to learn the weight of each basis function; while non-parametric approaches~\cite{Choi2008, Xu2011, Luo2018, Luo2019,Todescato2017,Benevento2020} model $\phi$ as the realization of a Gaussian Process and make predictions by conditioning on observed values of $\phi$ sampled over the operating environment. Alternative approaches to adaptive coverage~\cite{Davison2015, Choi2010} have also been considered.  

In this paper, we focus on a non-parametric adaptive coverage algorithm with provable regret guarantees. Similar adaptive coverage algorithms with formal performance guarantees are also developed in \cite{Todescato2017, Benevento2020}. 
Todescato~\etal~\cite{Todescato2017} use a Bernoulli random variable for each robot to decide between learning and coverage steps. The distribution of this random variable is designed to ensure convergence of the algorithm. In contrast, we leverage the so-called ``doubling trick" from the bandits literature to design a deterministic schedule of learning and coverage. This allows us to derive formal regret bounds on our adaptive coverage algorithm. 

Benevento~\etal~\cite{Benevento2020} use a Gaussian process optimization~\cite{NS-AK-SMK-MS:12} based approach to design an adaptive coverage algorithm and derive an upper bound on the regret with respect to coverage cost. However, they make  a strong assumption that Lloyd's algorithm converges to the global minimum of coverage cost. In contrast, our regret is defined with respect to the local minima that the Lloyd's algorithm will achieve starting at the current configuration and consequently, the regret bounds do not require such assumption.




We propose an adaptive coverage algorithm that balances the exploration-exploitation trade-off in \emph{learning} $\phi$ and achieving environmental \emph{coverage}.  Our algorithm schedules learning and coverage epochs such that its emphasis gradually shifts from exploration to exploitation while never fully ceasing to learn. 
We discuss analytical properties of our algorithm, and show that it achieves sublinear regret. 

The major contributions of this work are threefold. First, we propose Deterministic Sequencing of Learning and Coverage (DSLC), a novel adaptive coverage algorithm designed to balance the aforementioned exploration-exploitation trade-off. Second, we introduce a novel coverage regret that characterizes the deviation of agent configurations and partitions from a centroidal Voronoi partition and derive analytic bounds on the expected cumulative regret for DSLC. In particular, we prove that DSLC will achieve sublinear expected cumulative regret  under minor assumptions.
%
Third, we illustrate the efficacy of DSLC through extensive simulation and comparison with existing state-of-the-art approaches to adaptive coverage.

The remainder of the paper is organized as follows. The problem setup and related mathematical preliminaries are presented in Section \ref{sec:problem}. 
The DSLC algorithm is presented and analyzed in Sections \ref{sec:algorithm} and \ref{sec:analysis}, respectively. 
The performance of DSLC is elucidated through empirical simulations and is compared it with the state-of-the-art algorithms in Section \ref{sec:simulations}. Conclusions and future directions are discussed in Section \ref{sec:conclusions}.


\section{Problem Formulation}
\label{sec:problem}
We consider a team of $N$ agents tasked with providing coverage to a finite set of points in an environment represented by an undirected graph. The team is required to navigate within the graph to learn an unknown sensory function while maintaining near optimal configuration. In this section, we present the preliminaries of the estimation and coverage problem.

\subsection{Graph Representation of Environment}
We consider a discrete 
environment modeled by an undirected graph $G = (V, E)$, where the vertex set $V$ contains the finite set of points to be covered and the edge set $E  \subseteq V \times V$ is the collection of physically adjacent pairs of vertices that can be reached from each other without passing through other vertices. Let the weight map $w: E \rightarrow \real_{>0}$ indicate the distance between connected vertices. We assume $G$ is connected. Following standard definition of weighted undirected graph, a path in $G$ is an ordered sequence of vertices where there exist an edge between consecutive vertices. The distance between vertices $v_i$ and $v_j$ in $G$, denoted by $d_G(v_i,v_j)$, is defined by the minimum of the sums of the weights in the paths between $v_i$ and $v_j$.


Suppose there exists an unknown sensory function $\phi : V \rt \real_{>0}$ that assigns a nonnegative weight to each vertex in $G$. Intuitively, $\phi(v_i)$ could represent the intensity of signal of interest such as brightness or column of sound. A robot at vertex $v_i$ is capable of measuring $\phi(v_i)$ by collecting a sample $y = \phi(v_i)+ \epsilon$, where $\epsilon \sim \mathcal{N}(0, \sigma^2) $ is an additive zero mean Gaussian noise.

\subsection{Nonparametric Estimation}
Let $\bvec{\phi}$ be a vector with the $i$-th entry $\phi(v_i)$, $i \in \until{|V|}$, where $|\cdot|$ denotes set cardinality.
We assume a multivariate Gaussian prior for $\bvec{\phi}$ such that $\bvec{\phi} \sim \mathcal{N}( \bvec{\mu}_0 ,\bvec{\Lambda}_0^{-1})$, where  $\bvec{\mu}_0$ is mean vector and $\bvec{\Lambda}_0$ is the inverse covariance matrix. Let $n_i(t)$ be the number of samples and $s_i(t)$ be the summation of sampling results from $v_i$ until time $t$. Then, the posterior distribution of $\bvec{\phi}$ at time $t$ is $ \mathcal{N} \big( \bvec{\mu}(t) ,\bvec{\Lambda}^{-1}(t)\big)$~\cite{SMK:93}, where
\begin{equation}\label{posterior}
	\begin{split}
	\bvec{\Lambda}(t) &= \bvec{\Lambda}_0 + \sum_{i=1}^{\abs{V}} \frac{n_i(t)}{\sigma^2} \bvec{e}_i \bvec{e}_i^{\mathrm{T}}\\
	\bvec{\mu}(t) & = \bvec{\Lambda}^{-1}(t) \Bigg( \bvec{\Lambda}_0 \bvec{\mu}_0 + \sum_{i = 1}^{\abs{V}}\bvec{e}_i s_i(t) \Bigg). \\
	\end{split}
\end{equation}
Here, $\bvec{e}_i$ is the standard unit vector with $i$-th entry equal to $1$, $n_i(t)$ the total number of samples collected from $v_i$ and $s_i(t)$ is the summation of sampling results at $v_i$.

\subsection{Voronoi Partition and Coverage Problem}
We define the $N$-partition of graph $G$ as a collection $P = \seqdef{P_i}{i=1}^N$ of $N$ nonempty subsets of $V$ such that $\cup_{i=1}^N P_i = V$ and $P_i \cap P_j =\emptyset$ for any $i \neq j$. $P$ is said to be connected if the subgraph induced by $P_i$ denoted by $G[P_i]$ is connected for each $i \in N$. $G[P_i]$ being induced subgraph means its vertex set is $P_i$ and its edge set includes all edges in $G$ whose both end vertices are included in $P_i$. 

The configuration of the robot team is a vector of $N$ vertices $\bvec{\eta} \in V^N$ occupied by the robot team, where the $i$-th entry $\eta_i$ corresponds to position of the $i$-th robot. The $i$-th robot is tasked to cover vertices in $P_i$. Then, the coverage cost corresponding to configuration $\bvec{\eta}$ and connected $N$-partition $P$ can be defined as
\begin{equation}
\label{eq:cost}
\mathcal{H}(\bvec{\eta}, P) = \sum_{i=1}^N \sum_{v'\in P_i} d_{G[P_i]}(\eta_i,v') \phi(v'). 
\end{equation}
In a coverage problem, the objective is to minimize this coverage cost by selecting appropriate configuration $\bvec{\eta}$ and connected $N$-partition $P$. However, how to efficiently find the optimal configuration-partition pair in a large graph with arbitrary sensory function $\phi$ remains an open problem. There are two intermediate result about the optimal selection of configuration or partition when the other is fixed~\cite{Durham2012}.

\subsubsection{Optimal Partition with Fixed Configuration}
For a fixed configuration $\bvec{\eta}$ with distinct entries, a optimal connected $N$-partition $P$ minimizing coverage cost is called Voronoi partition denoted by $\vor(\bvec\eta)$. Formally, for each $P_i \in \vor(\bvec
\eta)$ and any $v' \in P_i$, \[d_G(v',\eta_i) \leq d_G(v',\eta_j), \quad \forall j\in \until{N} .\] 

\subsubsection{Optimal Configuration with Fixed Partition}
For a fixed connected $N$-partition $P$, the centroid of the $j$-th partition $P_j \in P$ is defined by
\[c_i := \argmin_{v \in P_i} \sum_{v' \in P_i} d_{G[P_i]}(v, v') \phi(v’),\]
and the optimal configuration is to place one robot at the centroid of each $P_i \in P$. We denote the vector of centroid of $P$ by $\bvec{c}(P)$ with $c_i$ as its $i$-th element.

Building upon the above two properties, the classic Lloyd algorithm iteratively places the robot to the centroid of the current Voronoi partition and computes the new Voronoi partition using the updated configuration. It is known that the robot team eventually converge to a class of partition called centroidal Voronoi partition defined below.

\begin{definition}[Centroidal Voronoi Partition,~\cite{DistCtrlRobotNetw}]
An $N$-partition $P$ is a centroidal Voronoi partion of $G$ if $P$ is a Voronoi partition generated by some configuration with distinct entries $\bvec{\eta}$, i.e. $P = \mathcal{V}(\bvec{\eta})$, and 
$\bvec{c}\left(\mathcal{V}(\bvec{\eta})\right) = \bvec{\eta}.$
\end{definition}

It needs to be noted that an optimal partition and configuration pair minimizing the coverage cost $\mathcal{H}(\bvec{\eta}, P)$ is of the form $(\bvec{\eta}^*, \mathcal{V}(\bvec{\eta}^*))$, where $\bvec{\eta}^*$ has distinct entries and $\mathcal{V}(\bvec{\eta}^*)$ is a centroidal Voronoi partition. A configuration-partition pair $(\bvec{\eta}', \mathcal{V}(\bvec{\eta}'))$ is considered to be an efficient solution to the coverage problem if $\mathcal{V}(\bvec{\eta}')$ is a centroidal Voronoi partition, even though it is possibly suboptimal~\cite{DistCtrlRobotNetw}.

\subsection{Performance Evaluation}
To achieve efficient coverage, the agents 
need to balance the trade-off between sampling the environment to learn $\bvec{\phi}$ (exploration) and achieving centroidal Voronoi configuration defined using the estimated $\phi$ (exploitation).  To characterize this trade-off, we introduce a notion of coverage regret. 


\begin{definition}[Coverage Regret]\label{def:regret}
	At each time $t$, let the team configuration be $\bvec{\eta}_t$ and the connected $N$-partition be $P_t$. The coverage regret until time $T$ is defined by $\sum_{t=1}^T R_t(\phi)$, where $R_t(\phi)$ is the instantaneous coverage regret with respect to sensory function $\phi$, and is defined by
	\begin{align*}
	R_t(\phi) &=  2\mathcal{H}(\bvec{\eta}_t, P_t) - \mathcal{H}(\bvec{c}(P_t), P_t) - \mathcal{H}(\bvec{\eta}_t, \mathcal{V}(\bvec{\eta}_t)),
	\end{align*}
which is the sum of two terms $\mathcal{H}(\bvec{\eta}_t, P_t) - \mathcal{H}(\bvec{c}(P_t), P_t)$ and $\mathcal{H}(\bvec{\eta}_t, P_t) -  \mathcal{H}(\bvec{\eta}_t, \mathcal{V}(\bvec{\eta}_t))$. The former (resp., latter) term is the regret induced by the deviation of the current configuration (resp., partition) from the optimal configuration (resp., partition) for the current partition (resp., configuration). Accordingly, no regret is incurred at time $t$ if and only if $P_t$ is a centroidal Voronoi $N$-partition and  $\bvec{\eta}_t = \bvec{c}(P_t)$. 

\end{definition}
There are two sources contributing to the coverage regret. First, the estimation error in the sensory function $\phi$. Second, the deviation from centroidal Voronoi partition while sampling environment to learn $\phi$.





\section{Deterministic Sequencing of Learning and Coverage (DSLC) Algorithm}
\label{sec:algorithm}
In this section, we describe the DSLC algorithm (Algorithm \ref{algo:DSLC}). It operates with an epoch-wise structure and each epoch consists of an exploration (learning) phase and an exploitation (coverage) phase. The exploration phase comprises two sub-phases: estimation and information propagation. 


\begin{algorithm}[t]	
	\smallskip
	{\footnotesize  
		\SetKwInOut{Input}{  Input}
		\SetKwInOut{Set}{  Set}
		\SetKwInOut{Title}{Algorithm}
		\SetKwInOut{Require}{Require}
		\SetKwInOut{Output}{Output}
		
		\Input{Environment graph $G$, $\bvec{\mu}_0$ , $\bvec{\Lambda}_0$ \;}
		
		\Set{$\alpha \in (0,1)$ and $\beta>1$\;}
		
		\medskip
		
		\For{epoch $j=1,2,\hdots$}{
			\smallskip
			\emph{\textbf{Exploration phase: }}
			
			\nl The robot team sample at vertices in $V$ to make 
			\[\max_{i\in \until{\abs{V}}}\sigma_i^2 (t) \leq  \alpha^j \sigma_0^2. \]
			
			\emph{\textbf{Information propagation phase: }}
			
			\nl Each robot agent propagate its sampling result to the team.
			
			\nl Each robot update estimated sensory function $\hat{\phi}$.
			
			\smallskip
			
			\emph{\textbf{Coverage phase: }}
			
			\nl \For{$t_j=1,2,\ldots, \ceil{\beta^j}$}{\ Based on $\hat{\phi}$, follow pairwise partitioning rule to update robot team configuration and partion.}

		}
		\caption{DSLC}
		\label{algo:DSLC}
	}
\end{algorithm}

\subsection{Estimation Phase}

Let the $\sigma_i^2(t)$ be the marginal posterior variance of $\phi(v_i)$ at time $t$, i.e. the $i$-th diagonal entry of $\bvec{\Lambda}^{-1}(t)$. Suppose all the the marginal prior variance $\sigma_i^2(0)$ is bounded by $\sigma_0^2$. Within each epoch $j$, agents first determine the points to be sampled in order to reduce $\max_{i\in \until{\abs{V}}}\sigma_i^2 (t)$ below a threshold $\alpha^{j}\sigma_0^2$, where $\alpha \in (0,1)$ is a prespecified parameter.

Notice that the posterior covariance computed in~\eqref{posterior} depends only the number of samples at each vertex, and does not require the actual sampling results. Therefore, the sequence of sampling locations can be computed before physically visiting the locations. Leveraging the deterministic evolution of the covariance, we take a greedy sampling policy that repeatedly selects the point $v_{i_t}$ with maximum marginal posterior variance, i.e.,
\begin{align}
i_t = \argmax_{i \in \abs{V}} \,\sigma_i(t),
\end{align}
for $t \in \{\underline{t}_j,\ldots, \overline{t}_j \}$, where $\underline{t}_j$ and $\overline{t}_j$ are the starting and ending time of estimation phase in the $j$-th epoch. It has been shown that the greedy sampling policy is near-optimal to maximize the mutual information of the sampling results and sensory function $\phi$~\cite{Krause2012}.

Let the set of points to be sampled during epoch $j$ be $X^j$ and let $X^j_r = X^j \cap P_{\underline{t}_j, r}$ be the set of sampling points that belong to $P_{\underline{t}_j, r}$, the partition assigned to agent $r$ at time $\underline{t}_j$. 
Each agent $r$ computes a path through the sampling points in $X^j_r $  and collects noisy measurements from those points. 


\begin{remark}
	With $\bvec{\Lambda}_0$ as the common knowledge, the set of sampling points $X^j$ for each epoch $j$ can be computed independently by each robot following the greedy sampling policy. If the same tie breaking rule is followed, the computed $X^j$ and the number of samples at each sampling point  are the same for all agents.
\end{remark}

\subsection{Information Propagation Phase}

After the estimation phase, the sampling results from each agent needs to be passed to all the other agents. There are several mechanisms to accomplish this in a finite number of steps. For example, agents can communicate with their neighboring agents and use flooding algorithms~\cite{lim2001flooding} to relay their sampling results to every agent. Alternatively, the agents may be able to send their sampling results to a cloud and receive global estimates after a finite delay. Another possibility for the agents is to use finite time consensus protocols~\cite{wang2010finite} in the distributed inference algorithm discussed in~\cite{PL-VS-NEL:16a}. 



For any of the above mechanisms, the sampling results from the entire robot team can be propagated to each robot agent in finite time. Then, each agent has an identical posterior distribution $ \mathcal{N} \big( \bvec{\mu}(t) ,\bvec{\Lambda}^{-1}(t)\big)$ of $\bvec{\phi}$, and  $\bvec{\mu}(t)$ will be used as the estimate $\hat{\phi}$ of the sensory function.


\subsection{Coverage Phase}
After the estimation and information propagation phases, agents have the same estimate of the sensory function $\hat{\phi}$. The coverage phase involves no environmental sampling and its length 
is designed to grow exponentially with epochs, i.e., the number of time steps in the coverage phase of  the $j$-th epoch is $\ceil{\beta^j}$ for some $\beta >1$. We use a distributed coverage algorithm, proposed in \cite{Durham2012}, called pairwise partitioning  with the estimated sensory function $\hat{\phi}$. 

In an connected $N$-partition $P$, $P_i$ and $P_j$ is said to be adjacent if there exists a vertex pair $v\in P_i$ and $v'\in P_j$ such that there exist an edge in $E$ connecting $v$ and $v'$. At each time, a random pair of agents $(i,j)$, with $P_i$ and $P_j$ adjacent, compute an optimal  pair of vertices $(a^*,b^*)$ within $ {P_i \cup P_j}$ that minimize 
\[ \sum_{v' \in P_i \cup P_j} \hat{\phi}(v')  \min\left(d_{G[P_i \cup P_j]}(a,v'), d_{G[P_i \cup P_j]}(b,v')\right).\]
Then, agents $i$ and $j$ move to $a^*$ and $b^*$. Subsequently, $P_i$ and $P_j$ are updated to
\begin{align*}
	P_i &\leftarrow  \setdef{v \in P_i \cup P_j }{d_{G[P_i \cup P_j]}(\eta_i, v ) \leq d_{G[P_i \cup P_j]}(\eta_j, v ) }\\
	P_j &\leftarrow  \setdef{v \in P_i \cup P_j }{d_{G[P_i \cup P_j]}(\eta_i, v ) > d_{G[P_i \cup P_j]}(\eta_j, v ) }.
\end{align*}


\section{Analysis of DSLC Algorithm}
\label{sec:analysis}
In this section, we analyze DSLC to provide an performance guarantee about the expected cumulative coverage regret. To this end, we leverage the information gain from the estimation phase to analyze the convergence rate of uncertainty. Then, we recall convergence properties of  the pairwise partitioning algorithm used in DSLC. Based on these results, we establish the main result of this paper, i.e., an upper bound on the expected cumulative coverage regret.

\subsection{Mutual Information and Uncertainty Reduction}

Let $X^g = (v_{i_1},\ldots,v_{i_n})$ be a sequence of $n$ vertices selected by the greedy policy and $Y_{X^g} = (y_1,\ldots, y_n)$ be observed sampling results corresponding to $X^g$. With a slight abuse of notation, we denote the marginal posterior variance of $\phi(v_i)$ after sampling at $v_{i_1} \ldots v_{i_k}$ by $\sigma^2_i(k)$. With greedy sampling policy,
\[i_k= \argmax_{i \in \until{\abs{V}}}\sigma_{i}^2(k-1).\]
Then, the mutual information of $Y^g$ and $\bvec{\phi}$ is
\begin{align}
I \left(Y_{X^g}; \bvec{\phi}  \right) &=  H\left( Y_{X^g}\right) - H\left( Y_{X^g} \,|\,\bvec{\phi}\right) \nonumber \\
&= \frac{1}{2} \sum_{k=1}^{n} \log \big( 1 + \sigma^{-2}\sigma_{i_k}^2\left(k-1\right) \big).\label{eq:mutual-info}
\end{align}
Let $\gamma_n$ be the maximal mutual information gain that can be achieved with $n$ samples. Then, 
\[
\gamma_n = \max_{X \in V^n } I\left(Y_X ; \bvec{\phi}\right).
\]
It is shown in~\cite{nemhauser1978analysis} that the  mutual information gain~\eqref{eq:mutual-info} achieved by the greedy sampling policy is near optimal, i.e., 
\begin{equation}\label{ineq:no}
	\left(1-\frac{1}{e}\right) \gamma_n \leq I \left(Y_{X_g}; \bvec{\phi} \right) \leq \gamma_n.
\end{equation}

We now present a bound on the maximal posterior variance after sampling at vertices within $X^g$. The following Lemma and proof techniques are adapted from our previous work~\cite{wei2020expedited} to incorporate the discrete environment.

\begin{lemma}[Uncertainty reduction]	\label{lemma: ur}
	 Under the greedy sampling policy, the maximum posterior variance after $n$ sampling rounds satisfies
	\begin{equation*}
	\max_{i \in \until{\abs{V}}} \sigma_i^2 (n) \leq \frac{2 \sigma_0^2 }{\log \left( 1 + \sigma^{-2}\sigma_0^2 \right)} \frac{\gamma_n}{n}.
	\end{equation*}
\end{lemma}

\smallskip 

\begin{proof}
	For any $i \in \until{\abs{V}}$, $\sigma_{i}^2(k)$ is monotonically non-increasing in $k$. So, we get
	\begin{equation}\label{ineq: variance}
	\begin{split}
	\sigma_{i_{k+1}}^2(k) &\leq \sigma_{i_{k+1}}^2(k-1) \\
	&\leq\max_{i \in \until{\abs{V}}}\sigma_{i}^2(k-1) =  \sigma_{i_k}^2(k-1),
	\end{split}	
	\end{equation} 
	which indicates that $\sigma_{i_{k+1}}^2(k)$ is monotonically non-increasing. Hence, from~\eqref{eq:mutual-info} and~\eqref{ineq:no}, $\log \left( 1 + \sigma^{-2}\sigma_{i_n}^2\left(n-1\right) \right)  \leq 2\gamma_n/n$.
	Since ${x^2}/{\log \left(1+x^2\right)}$ is an increasing function on $[0,\infty)$,
	\[{\sigma_{i_n}^2\left(n-1\right)} \leq \frac{\sigma_0^2}{\log \left( 1 + \sigma^{-2}\sigma_0^2 \right)} {\log \left( 1 + \sigma^{-2}\sigma_{i_n}^2\left(n-1\right) \right)}.\]
	Substituting~\eqref{ineq: variance} into the above equation, we conclude that
	\begin{align*}
	 {\sigma_{i_n}^2\left(n-1\right)}  \leq \frac{2 \sigma_0^2 }{\log \left( 1 + \sigma^{-2} \sigma_0^2  \right)} \frac{\gamma_n}{n},
	\end{align*}
	which establishes the lemma.
\end{proof}

Typically, it is hard to characterize $\gamma_n$ for with a general $\bvec{\Sigma}_0$. Therefore, we make the following assumption.

\begin{assumption}\label{assum: ig}
	Vertices in $V$ lie in a convex and compact set $D \in \real^2$ and  the covariance of any pair $\phi(v_i)$ and $\phi(v_j)$ is determined by an exponential kernel function 
	\begin{equation}\label{def: stker}
	k(\phi(v_i), \phi(v_j)) = \sigma_v^2 \exp\left(-\frac{\subscr{d}{eu}^2(v_i,v_j)}{2 l^2}\right),
	\end{equation}
	where $\subscr{d}{eu}(v_i,v_j)$ is the Euclidean distance between $v_i$ and $v_j$, $l$ is the length scale, and $\sigma_v^2$ is the variability parameter. 
\end{assumption} 

 We now recall an upper bound on $\gamma_n$ from~\cite{NS-AK-SMK-MS:12}.
\begin{lemma}[{Information gain for squared exp. kernel}] \label{lemma: parmi}
	With Assumption \ref{assum: ig}, the maximum mutual information satisfies $\gamma_n\in O( (\log \abs{V}n)^3)$.
\end{lemma}

\begin{remark}

If the correlation information is ignored, i.e., $\phi(i)$, $i \in \until{\abs{V}}$ are treated to be independent, it can be seen that $\max_{i \in \until{\abs{V}}} \sigma_i^2 (n) \in O (\abs{V}/n)$ with greedy sampling policy. In contrast, if correlation information is considered, by substituting the result in Lemma~\ref{lemma: parmi} into Lemma~\ref{lemma: ur}, $\max_{i \in \until{\abs{V}}} \sigma_i^2 (n) \in O ((\log (\abs{V}n))^3/n)$, which shows great advantage about reducing uncertainty when $\abs{V}$ is large (the environment is finely discretized).

\end{remark}

\subsection{Convergence within Coverage Phase}
Before each coverage phase, since the sampling results of each agent are relayed to the entire team, the team have a consensus estimate of the sensory function $\hat{\phi}$. It has been shown in~\cite{Durham2012} that using the pairwise partitioning algorithm, the $N$-partition $P$ for the team  converges almost surely to a class of near optimal partitions defined below.
\begin{definition}[Pairwise-optimal Partition]
	A connected $N$-partition $P$ is pairwise-optimal if for each pair of adjacent regions $P_i$ and $P_j$,
	\begin{align*}
		&\sum_{v'\in P_i} d_G(c(P_i),v') \phi(v') + \sum_{v'\in P_j} d_G(c(P_j),v') \phi(v') \\
		=& \min_{a,b \in P_i \cup P_j} \sum_{v' \in P_i \cup P_j} \phi(v')  \min\left(d_G(a,v'), d(b,v')\right).
	\end{align*}
\end{definition}
It means that, within the induced subgraph generated by any pair of adjacent regions, the $2$-partition is optimal. It is proved in~\cite{Durham2012} that if a connected $N$-partition $P$ is pairwise-optimal then it is also a centroidal Voronoi partition. The following result on the convergence time of pairwise partitioning algorithm is established in~\cite{Durham2012}. 
\begin{lemma}[Expected Convergence Time] \label{lemma: ect}
	Using the pairwise partitioning algorithm, the expected time to converge to a pairwise-optimal $N$-Partition is finite. 
\end{lemma}

\smallskip 

For each coverage phase, Lemma~\ref{lemma: ect} implies that the expected time for the instantaneous regret $R_t(\hat{\phi})$ to converge to  $0$ is finite.

\subsection{An Upper Bound on Expected Coverage Regret}
We now present the main result for this paper.

\begin{theorem}\label{theorem:regret}
	For DSLC and any time horizon $T$, if Assumption~\ref{assum: ig} holds and $\alpha= \beta^{-2/3}$, then the expected cumulative coverage regret with respect sensory function $\phi$ satisfies
	\[ \expt \Bigg[ \sum_{t=1}^{T} R_t(\phi)\Bigg]\in O \big (T^{2/3} (\log(T))^3 \big).\]
\end{theorem}

\medskip

\begin{proof}
	We establish the theorem using the following four steps. 
	
	\noindent \textbf{Step 1 (Regret from estimation phases):}
	Let the total number of sampling steps before the end of the $j$-th epoch be $s_j$. By applying Lemma~\ref{lemma: ur}, we get 
	\[s_j  \in  O ({(\log(T))^3}/{\alpha^j}).\]
Thus, the coverage regret in the estimation phases until the end of the $j$-th epoch  belongs to $O ({(\log(T))^3}/{\alpha^j})$. 
	
	\noindent \textbf{Step 2 (Regret from information propagation phases):}
	The sampling information by each robot propagate to all the team members in finite time. Thus, before the end of the $j$-th epoch, the coverage regret from information propagation phases can be bounded by $c_1 j$ for some constant $c_1>0$.	
	
	\noindent \textbf{Step 3 (Regret from coverage phases):}
	According to Lemma~\ref{lemma: ect}, in each coverage phase, the expected time before converging to a pairwise-optimal partition is finite. Thus, before the end of the $j$-th epoch, the expected coverage regret from converging steps can be upper bounded by $c_2 j$ for some constant $c_2>0$.
	
	Also note that the robot team converge to pair-wise optimal partition with respect estimated sensory function $\hat{\phi}$ which may deviate from the actual $\phi$. Then,  the instantaneous coverage regret $R_t(\phi)$ caused by estimation error can be expressed as
	\[2\mathcal{H}(\bvec{\eta}_t, P_t) - \mathcal{H}(\bvec{c}(P_t), P_t) - \mathcal{H}(\bvec{\eta}_t\mathcal{V}(\bvec{\eta}_t)) := A_t^{\mathrm{T}} \bvec{\phi}, \]
	for some $A_t \in \real^{\abs{V}}$. Moreover, the posterior distribution of $R_t(\phi)$ can be written as $\mathcal{N} (A_t^{\mathrm{T}} \bvec{\mu}(t), A_t^{\mathrm{T}} \bvec{\Sigma}(t) A_t) $, where $\bvec{\Sigma}(t) = \bvec{\Lambda}^{-1}(t)$ is the posterior covariance matrix. Since a pairwise-optimal partition $P$ is also a centroidal Voronoi partition and $\hat{\phi} = \bvec{\mu}(t)$,  $R_t(\hat{\phi}) = 0 $ indicates $A_t^{\mathrm{T}} \bvec{\mu}(t) = 0$. 
	Now, we get $R_t(\phi) \sim \mathcal{N} (0, A_t^{\mathrm{T}} \bvec{\Sigma}(t) A_t)$ and
	\[\expt[R_t(\phi)] \leq \expt\left[\abs{R_t(\phi)}\right] =  \sqrt{\frac{2}{\pi} A_t^{\mathrm{T}} \bvec{\Sigma}(t) A_t}. \]
	Note that $A_t^{\mathrm{T}} \bvec{\Sigma}(t) A_t$ is weighed summation of eigenvalues of $\bvec{\Sigma}(t)$. At any time $t$ in the coverage phase of the $k$-th epoch, $\max_{i\in \until{\abs{V}}}\sigma_i^2 (t) \leq \alpha^{k}\sigma_0^2$, and its follows that the summation of eigenvalues of $\bvec{\Sigma}(t)$ equals $\text{trace}(\bvec{\Sigma}(t)) \leq \abs{V}\alpha^{k}\sigma_0^2$. Thus, we get 
	\[\expt \Bigg [\sum_{t\in \supscr{\mc T_k}{cov}}  R_t(\phi) \Bigg] \leq c_3 (\beta \sqrt{\alpha})^{k},  \]
	for some constant $c_3>0$, where $\supscr{\mc T_k}{cov}$ are the time slots in the coverage phase of the $k$-th epoch and we have used the fact that $|\supscr{\mc T_k}{cov}| = \lceil \beta^k \rceil$.

	\noindent \textbf{Step 4 (Summary):} Summing up the expected coverage regret from above steps, the expected cumulative coverage regret at the end of the $j$-th epoch $T_j$ satisfies
	\begin{align}
		\expt \Bigg[ \sum_{t=1}^{T_j} R_t(\phi)\Bigg] \leq C_1 j +C_2 s_j +  \sum_{k=1}^j c_3 (\beta\sqrt{\alpha})^{k}, 
	\end{align}
	where $C_1, C_2 >0$ are some constants. 
	The theorem statement follows by plugging in $\alpha= \beta^{-2/3}$, using $j \in O(\log T)$ and some simple calculations. 
\end{proof}



\section{Simulation Results}
\label{sec:simulations}
To illustrate the empirical performance of the proposed algorithm, we simulate its execution on a uniform grid graph superimposed on the unit square. We present numerical results which show that DSLC achieves sublinear regret and compare our algorithm to those proposed in \cite{Cortes2004} and \cite{Todescato2017}.

\begin{figure}
    \centering
    \includegraphics[width=0.5\textwidth]{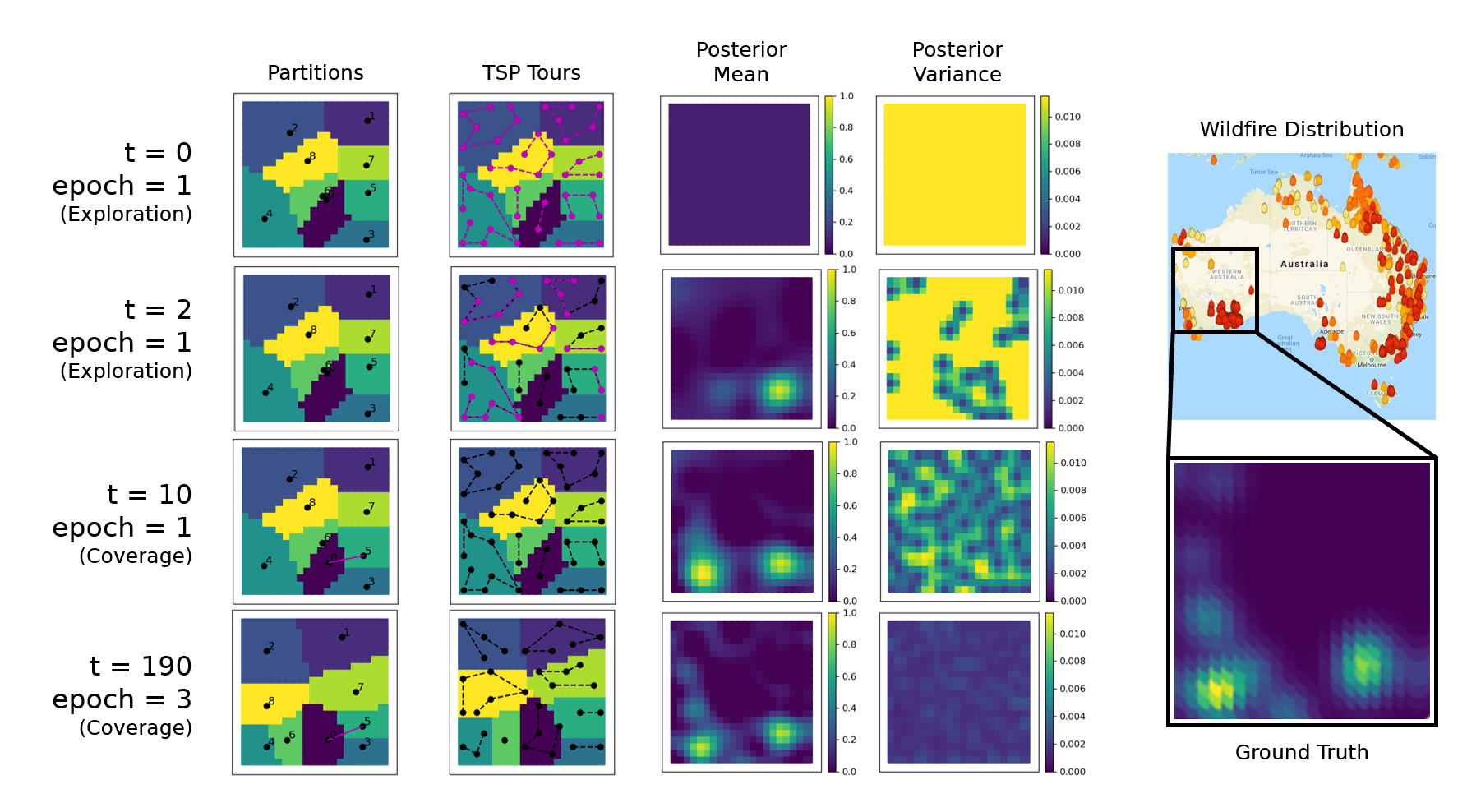}
    \caption{
    Distributed implementation of DSLC in the unit square with $j=3$ epochs of length 16, 46, and 128.
    From left to right: agent positions $\bvec{\eta}_t$ and partitions $P_t$; TSP sampling tours; posterior mean and variance of $\hat{\phi}$; ground truth sensory function $\phi$ based on data from \cite{Paradis2019}. Pairwise partition updates between gossiping agents are denoted by magenta lines in the leftmost column of panels. Points along TSP tours in second-from-leftmost column of panels are plotted in magenta prior to sampling, and in black after sampling. Video is available online.\textsuperscript{1}
    }
    \vspace{-1.5em}
    \label{fig:simulation_gossip}
\end{figure}

Motivated by environmental applications, we construct the sensory function $\phi$ over a discrete 21 $\times$ 21 point grid in $[0, 1]^2$ by performing kernel density estimation on a subset of the geospatial distribution of Australian wildfires observed by NASA in 2019 \cite{Paradis2019}. Intuitively, $\phi$ represents the probability that a wildfire may occur at a particular point of the unit square, and may be used to model the demand for an autonomous sensing agent at that point. ``Hotspots'' in which wildfires are highly likely correspond to areas of high demand for sensing agents, while areas in which few wildfires have occurred are assumed to correspond to areas of low demand for agents. The ground truth $\phi$ obtained through kernel density estimation is shown on the right in Figure \ref{fig:simulation_gossip}.


In each simulation, nine agents are placed uniformly at random over the grid and execute three epochs of length 16, 46, and 128 to achieve adaptive coverage of the environment. Partitions are initialized by iterating over the grid and assigning each point to the nearest agent. During the exploration phase of each epoch, partitions are fixed; during the exploitation phase of each epoch, partitions are updated according to the protocol established in \cite{Durham2012}, where pairwise gossip-based repartitioning occurs between randomly selected neighbors. Coverage cost, regret and maximum variance are computed throughout using \eqref{eq:cost}, Definition \ref{def:regret}, and the maximum diagonal entry of $\bvec{\Lambda}^{-1}(t)$ from \eqref{posterior}, respectively.

The sensory function $\phi$ is normalized in the range $[0, 1]$ and sampled by agents with Gaussian noise parameterized by mean and standard deviation $\mu=0, \; \sigma=0.1$. A global Gaussian Process model is assumed to simplify estimation of $\hat{\phi}$ throughout the simulation, though in practice estimation of $\hat{\phi}$ could be implemented in a fully-distributed manner by assuming each agent maintains its own model of $\hat{\phi}$ and employing an information propagation phase described in Section \ref{sec:algorithm}. Setting the parameter $\alpha=0.5$ to reduce uncertainty by half within each epoch, $\beta = \alpha^{-3/2}$ is fully determined by Theorem \ref{theorem:regret}. Figure \ref{fig:simulation_gossip} visualizes the simulation of DSLC. A video of the simulation is available online.\footnote{\texttt{https://youtu.be/nalwrZC6GiI}}

\begin{figure*}
    \centering
    \includegraphics[width=0.9\textwidth]{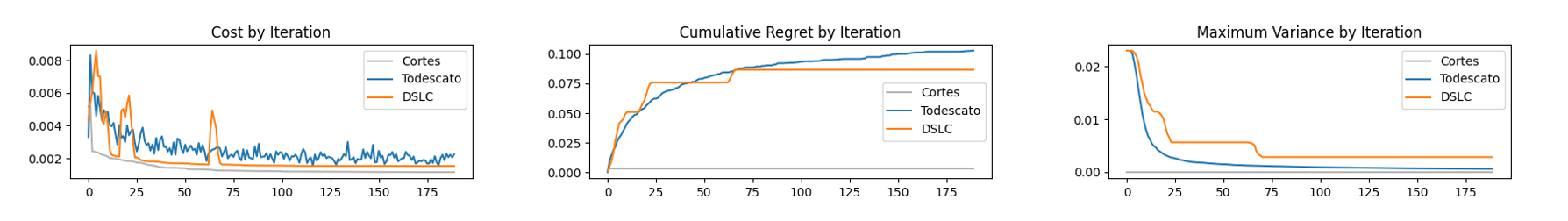}
    \caption{Cost \eqref{eq:cost}, regret (Definition \ref{def:regret}) and maximum posterior variance \eqref{posterior} of $\hat{\phi}$ for DSLC, \texttt{Todescato} and \texttt{Cortes} averaged over 16 simulations of 190 iterations each. Note that DSLC empirically achieves sublinear regret. Spikes in regret occur during the exploration phase of each epoch, before agents converge to a pairwise-optimal coverage configuration with respect to $\hat{\phi}$ during the exploitation phase.
    }
    \vspace{-1em}
    \label{fig:regret}
\end{figure*}

Figure \ref{fig:regret} compares the evolution of the regret and cost in DSLC with that of algorithms proposed in \cite{Cortes2004} and \cite{Todescato2017}, denoted \texttt{Cortes} and \texttt{Todescato}, respectively. As described in Section \ref{sec:introduction}, agents in \cite{Cortes2004} are assumed to have perfect knowledge of $\phi$ and simply go to the centroid
of their cell at each iteration; in \cite{Todescato2017}, agents follow a stochastic sampling approach with probability of exploration proportional to posterior variance in the estimate $\hat{\phi}$ at each iteration. All results are averaged over 16 simulations of 190 iterations, aligned with the  three-implementation of DSLC with epoch lengths 16, 46, and 128.

Though we do not include the algorithm in our simulations, it is worth noting that DSLC operates in a manner similar to that proposed in \cite{Choi2008} where agents spend a number of iterations sampling $\phi$ to reduce maximum posterior variance $\max_{i \in \until{\abs{V}}} \sigma_i^2 (n)$ below a prespecified threshold, then transition to perform coverage for all remaining iterations. Indeed, this algorithm is essentially a special case of DSLC with one epoch and can therefore be expected to perform similarly from an empirical perspective.


\section{Conclusions}
\label{sec:conclusions}
In this paper, we study the distributed multi-robot coverage problem over an unknown nonuniform sensory field. DSLC, a novel adaptive coverage algorithm designed to drive agents to simultaneously learn the sensory function and provide satisfactory coverage, is proposed. Defining a novel characterization of coverage regret, we analyze DSLC and bound its coverage regret as sublinear. We illustrate the empirical promise of DSLC through simulations in which a team of aerial robots is tasked with coverage of an unknown geospatial distribution of wildfires. 

In future works, we hope to extend our approach to settings in which agents are assumed to have heterogeneous sensing and motion capabilities. We also see potential for extension to nonstationary settings in which a sensing field evolves with time. Such settings more accurately reflect the challenges presented by real-world implementation of multi-robot control algorithms, and offer promising avenues to broader impacts in multi-robot systems research. 




 
\footnotesize 

\bibliographystyle{IEEEtran}
\balance
\bibliography{main_arxiv}

\begin{thebibliography}{10}
\providecommand{\url}[1]{#1}
\csname url@samestyle\endcsname
\providecommand{\newblock}{\relax}
\providecommand{\bibinfo}[2]{#2}
\providecommand{\BIBentrySTDinterwordspacing}{\spaceskip=0pt\relax}
\providecommand{\BIBentryALTinterwordstretchfactor}{4}
\providecommand{\BIBentryALTinterwordspacing}{\spaceskip=\fontdimen2\font plus
\BIBentryALTinterwordstretchfactor\fontdimen3\font minus
  \fontdimen4\font\relax}
\providecommand{\BIBforeignlanguage}[2]{{%
\expandafter\ifx\csname l@#1\endcsname\relax
\typeout{** WARNING: IEEEtran.bst: No hyphenation pattern has been}%
\typeout{** loaded for the language `#1'. Using the pattern for}%
\typeout{** the default language instead.}%
\else
\language=\csname l@#1\endcsname
\fi
#2}}
\providecommand{\BIBdecl}{\relax}
\BIBdecl

\bibitem{Cortes2004}
J.~Cort{\'{e}}s, S.~Mart{\'{i}}nez, T.~Karataş, and F.~Bullo, ``{Coverage
  control for mobile sensing networks},'' \emph{IEEE Transactions on Robotics
  and Automation}, vol.~20, no.~2, pp. 243--255, 2004.

\bibitem{Cortes2005}
J.~Cort{\'{e}}s and F.~Bullo, ``{Coordination and Geometric Optimization via
  Distributed Dynamical Systems},'' \emph{SIAM Journal on Control and
  Optimization}, vol.~44, no.~5, pp. 1543--1574, 2005.

\bibitem{Lekien2010}
F.~Lekien and N.~E. Leonard, ``{Nonuniform coverage and cartograms},''
  \emph{Proceedings of the IEEE Conference on Decision and Control}, pp.
  5518--5523, 2010.

\bibitem{hussein2007effective}
I.~I. Hussein and D.~M. Stipanovic, ``Effective coverage control for mobile
  sensor networks with guaranteed collision avoidance,'' \emph{IEEE
  Transactions on Control Systems Technology}, vol.~15, no.~4, pp. 642--657,
  2007.

\bibitem{Lloyd1982}
S.~P. Lloyd, ``{Least Squares Quantization in PCM},'' \emph{IEEE Transactions
  on Information Theory}, vol.~28, no.~2, pp. 129--137, 1982.

\bibitem{Bullo2012}
F.~Bullo, R.~Carli, and P.~Frasca, ``{Gossip coverage control for robotic
  networks: Dynamical systems on the space of partitions},'' \emph{SIAM Journal
  on Control and Optimization}, vol.~50, no.~1, pp. 419--447, 2012.

\bibitem{Durham2012}
J.~W. Durham, R.~Carli, P.~Frasca, and F.~Bullo, ``{Discrete partitioning and
  coverage control for gossiping robots},'' \emph{IEEE Transactions on
  Robotics}, vol.~28, no.~2, pp. 364--378, 2012.

\bibitem{Schwager2009}
M.~Schwager, D.~Rus, and J.~J. Slotine, ``{Decentralized, adaptive coverage
  control for networked robots},'' \emph{International Journal of Robotics
  Research}, vol.~28, no.~3, pp. 357--375, 2009.

\bibitem{Schwager2017}
M.~Schwager, M.~P. Vitus, S.~Powers, D.~Rus, and C.~J. Tomlin, ``{Robust
  adaptive coverage control for robotic sensor networks},'' \emph{IEEE
  Transactions on Control of Network Systems}, vol.~4, no.~3, pp. 462--476,
  2017.

\bibitem{Choi2008}
J.~Choi, J.~Lee, and S.~Oh, ``{Swarm intelligence for achieving the global
  maximum using spatio-temporal Gaussian processes},'' \emph{Proceedings of the
  American Control Conference}, pp. 135--140, 2008.

\bibitem{Xu2011}
Y.~Xu and J.~Choi, ``{Adaptive sampling for learning Gaussian processes using
  mobile sensor networks},'' \emph{Sensors}, vol.~11, no.~3, pp. 3051--3066,
  2011.

\bibitem{Luo2018}
W.~Luo and K.~Sycara, ``{Adaptive Sampling and Online Learning in Multi-Robot
  Sensor Coverage with Mixture of Gaussian Processes},'' \emph{Proceedings -
  IEEE International Conference on Robotics and Automation}, pp. 6359--6364,
  2018.

\bibitem{Luo2019}
W.~Luo, C.~Nam, G.~Kantor, and K.~Sycara, ``{Distributed environmental modeling
  and adaptive sampling for multi-robot sensor coverage},'' in
  \emph{Proceedings of the International Joint Conference on Autonomous Agents
  and Multiagent Systems, {AAMAS}}, 2019, pp. 1488--1496.

\bibitem{Todescato2017}
M.~Todescato, A.~Carron, R.~Carli, G.~Pillonetto, and L.~Schenato,
  ``{Multi-robots Gaussian estimation and coverage control: From
  client–server to peer-to-peer architectures},'' \emph{Automatica}, vol.~80,
  pp. 284--294, 2017.

\bibitem{Benevento2020}
A.~Benevento, M.~Santos, G.~Notarstefano, K.~Paynabar, M.~Bloch, and
  M.~Egerstedt, ``Multi-robot coordination for estimation and coverage of
  unknown spatial fields,'' in \emph{IEEE International Conference on Robotics
  and Automation (ICRA)}, 2020, pp. 7740--7746.

\bibitem{Davison2015}
P.~Davison, N.~E. Leonard, A.~Olshevsky, and M.~Schwemmer, ``{Nonuniform Line
  Coverage from Noisy Scalar Measurements},'' \emph{IEEE Transactions on
  Automatic Control}, vol.~60, no.~7, pp. 1975--1980, 2015.

\bibitem{Choi2010}
J.~Choi and R.~Horowitz, ``{Learning coverage control of mobile sensing agents
  in one-dimensional stochastic environments},'' \emph{IEEE Transactions on
  Automatic Control}, vol.~55, no.~3, pp. 804--809, 2010.

\bibitem{NS-AK-SMK-MS:12}
N.~Srinivas, A.~Krause, S.~M. Kakade, and M.~Seeger, ``Information-theoretic
  regret bounds for {G}aussian process optimization in the bandit setting,''
  \emph{IEEE Transactions on Information Theory}, vol.~58, no.~5, pp.
  3250--3265, 2012.

\bibitem{SMK:93}
S.~M. Kay, \emph{Fundamentals of Statistical Signal Processing, Volume I :
  Estimation Theory}.\hskip 1em plus 0.5em minus 0.4em\relax Prentice Hall,
  1993.

\bibitem{DistCtrlRobotNetw}
F.~Bullo, J.~Cort\'es, and S.~Mart{\'\i}nez, \emph{Distributed Control of
  Robotic Networks}, ser. Applied Mathematics Series.\hskip 1em plus 0.5em
  minus 0.4em\relax Princeton University Press, 2009, electronically available
  at http://coordinationbook.info.

\bibitem{Krause2012}
A.~Krause and C.~E. Guestrin, ``Near-optimal nonmyopic value of information in
  graphical models,'' in \emph{Conf.\ on Uncertainty in Artificial
  Intelligence}, Edinburgh, Scotland, Jul. 2005, pp. 324--331.

\bibitem{lim2001flooding}
H.~Lim and C.~Kim, ``Flooding in wireless ad hoc networks,'' \emph{Computer
  Communications}, vol.~24, no. 3-4, pp. 353--363, 2001.

\bibitem{wang2010finite}
L.~Wang and F.~Xiao, ``Finite-time consensus problems for networks of dynamic
  agents,'' \emph{IEEE Transactions on Automatic Control}, vol.~55, no.~4, pp.
  950--955, 2010.

\bibitem{PL-VS-NEL:16a}
P.~Landgren, V.~Srivastava, and N.~E. Leonard, ``Distributed cooperative
  decision-making in multiarmed bandits: Frequentist and {B}ayesian
  algorithms,'' in \emph{IEEE Conf. on Decision and Control}, Las Vegas, NV,
  Dec. 2016, pp. 167--172.

\bibitem{nemhauser1978analysis}
G.~L. Nemhauser, L.~A. Wolsey, and M.~L. Fisher, ``An analysis of
  approximations for maximizing submodular set functions,'' \emph{Mathematical
  programming}, vol.~14, no.~1, pp. 265--294, 1978.

\bibitem{wei2020expedited}
L.~Wei, X.~Tan, and V.~Srivastava, ``Expedited multi-target search with
  guaranteed performance via multi-fidelity {G}aussian processes,'' in
  \emph{IEEE/RSJ International Conference on Intelligent Robots and Systems},
  Las Vegas, NV (Virtual), Oct. 2020, pp. 7095--7100.

\bibitem{Paradis2019}
\BIBentryALTinterwordspacing
C.~Paradis, ``{Fires from Space: Australia},'' 2019. [Online]. Available:
  \url{https://www.kaggle.com/carlosparadis/fires-from-space-australia-and-new-zeland}
\BIBentrySTDinterwordspacing

\end{thebibliography}

\end{document}